\documentclass[3p,,,12pt]{elsarticle}
\usepackage{hyperref}

\journal{arXiv}

\begin{document}

\begin{frontmatter}

\title{Untargeted Region of Interest Selection for GC-MS Data using a Pseudo F-Ratio Moving Window ($\psi$FRMV)}

\author[mymainaddress]{Ryland T. Giebelhaus}
\author[mymainaddress]{Michael D. Sorochan Armstrong}
\author[mymainaddress]{A. Paulina de la Mata}
\author[mymainaddress]{James J. Harynuk\corref{mycorrespondingauthor}}

\cortext[mycorrespondingauthor]{Corresponding author}
\ead{james.harynuk@ualberta.ca}

\address[mymainaddress]{Department of Chemistry, University of Alberta, 11227 Saskatchewan Dr NW, Edmonton, Alberta, Canada}

\begin{abstract}
 There are many challenges associated with analysing gas chromatography - mass spectrometry (GC-MS) data. Many of these challenges stem from the fact that electron ionisation (EI) can make it difficult to recover molecular information due to the high degree of fragmentation with concomitant loss of molecular ion signal. With GC-MS data there are often many common fragment ions shared among closely-eluting peaks, necessitating sophisticated methods for analysis. Some of these methods are fully automated, but make some assumptions about the data which can introduce artifacts during the analysis. Chemometric methods such as Multivariate Curve Resolution (MCR), or Parallel Factor Analysis (PARAFAC/PARAFAC2) are particularly attractive, since they are flexible and make relatively few assumptions about the data - ideally resulting in fewer artifacts. These methods do require expert user intervention to determine the most relevant regions of interest and an appropriate number of components, $k$, for each region. Automated region of interest selection is needed to permit automated batch processing of chromatographic data with advanced signal deconvolution. Here, we propose a new method for automated, untargeted region of interest selection that accounts for the multivariate information present in GC-MS data to select regions of interest based on the ratio of the squared first, and second singular values from the Singular Value Decomposition (SVD) of a window that moves across the chromatogram. Assuming that the first singular value accounts largely for signal, and that the second singular value accounts largely for noise, it is possible to interpret the relationship between these two values as a probabilistic distribution of Fisher Ratios. The sensitivity of the algorithm was tested by investigating the concentration at which the algorithm can no longer pick out chromatographic regions known to contain signal. The algorithm achieved detection of features in a GC-MS chromatogram at concentrations below 10 pg on-column. The resultant probabilities can be interpreted as regions that contain features of interest.
\end{abstract}


\end{frontmatter}


\section{Introduction}

Gas Chromatography - Mass Spectrometry (GC-MS) is a mature analytical technique that offers profound insight into the volatile and/or semi-volatile chemical characteristics of a variety of different samples. In recent years, owing to the increased interest in -omics type research, GC-MS has been routinely applied to challenging biological samples including metabolite profiling in plants and bodily fluids such as urine \cite{fiehn2000metabolite, fiehn2002metabolomics, bouatra2013human}. Because of the chemical diversity associated with these and other complex samples such as petroleum, the need for accurate and representative analysis of GC-MS data has remained a topical research problem \cite{tran2006comparison}. 

A number of different techniques for the analysis of GC-MS data have been proposed. Many of these techniques have been deployed as part of a commercial suite of software, although these packages tend to be poorly transparent and often rely on many different user-selected parameters to extract the chemical information from the raw signal \cite{lu2008comparative}. This introduces an element of subjectivity to the analysis, in addition to the fact that commercial software is almost always compiled and difficult to interact with or build upon.

Chemometric methods such as Multivariate Curve Resolution (MCR), Independent Component Analysis (ICA), or Parallel Factor Analysis (PARAFAC/PARAFAC2) for the decomposition of multiple GC-MS samples simultaneously, present a number of advantages over various commercial offerings \cite{van2009global, domingo2016automated, amigo2008solving}. Advances using these technologies have been well documented in the literature and their limitations have also been thoroughly explored, which allows for a more productive discussion on the limitations of the instrumentation versus the limitations of the data analysis tools.

To minimize the phenomenon of ``peak drop-out'', where chemical factors are not identified across all samples in which they are present, a popular trend appears to be application of PARAFAC2 to handle GC-MS data, as it can account for drift along the chromatographic mode\cite{johnsen2017gas}. Using PARAFAC2, the presence of each component is solved for across each component via a series of regression steps, which are informed by multiple samples. With a properly optimised PARAFAC2 model, the analyst can be assured that the presence or absence of a chemical component in a particular region is not due to the failure of a dynamic programming routine that associates factors that were integrated and deconvolved separately. PARAFAC2, while certainly a powerful way of analysing GC-MS data, does require pre-defined regions of interest within each of the chromatograms, and an appropriate $k$-component number for each region. These are difficult determinations to automate \cite{amigo2008solving}. Despite this, a fully automated method for the deployment of PARAFAC2 for a GC-MS data-set has been recently applied using a deep-learning routine, which demonstrates that a fully-automated method is possible \cite{baccolo2021untargeted}.

Region of interest (ROI) selection is a broad field in computer science and statistics, focusing on selecting regions in a data set that contain  useful (or potentially useful) information   \cite{poldrack2007region}. ROI selection serves three main purposes in data processing and pre-processing; it (1) allows for a more clear manual interpretation of the data by highlighting areas which are defined as significant or interesting by the algorithm, (2) controls for type 1 error by limiting the number of statistical tests to a few ROIs, and (3) limits testing and analysis to a set region defined by an automated ROI algorithm \cite{poldrack2007region, baccolo2021untargeted, 5076670}. Automated ROI selection is popular in signal and image processing as it selects ROIs and provides an output based on a series of predefined metrics, eliminating bias that is frequently introduced by manual ROI selection \cite{5076670, poldrack2007region}. Here, the we present a novel method for identifying ROIs in GC-MS data for subsequent analysis. Using two relatively intuitive and robust parameters, it is possible to automate half of the remaining algorithmic work associated with PARAFAC-based decomposition of GC-MS data. As outputs, the algorithm returns a matrix containing the original GC-MS signals in regions of the data identified as ROIs. The portion of the data deemed to be outside of an ROI is zeroed. Thus, the original mass spectral signals, retention information and peak shapes are preserved. We tested the dynamic range and sensitivity of our algorithm by processing a series of chromatograms of GC standards across a wide range of concentrations; from 1000 pg on-column to 5 pg on-column. We also attempt to generalise some prior research relying on F-ratios for supervised classification of GC$\times$GC-TOFMS samples \cite{parsons2015tile,marney2013tile,pierce2015pixel}, in way that does not require \textit{a-priori} class information.

\section{Experimental}

\subsection{Region of interest selection algorithm}
Development and application of the region of interest algorithm was performed in Matlab$\textsuperscript{\textregistered}$ R2021a. Each chromatogram was imported into Matlab as a data matrix, $X \in \mathbf{R}^{I\times J}$, where \textit{I} is the number of acquisitions, and \textit{J} is the number of mass-to-charge ratios (\textit{m/z}) per spectra. Each $i, j$ entry describing the relative abundance of a particular fragment at a moment in time along a chromatogram.

\subsubsection{Algorithm inputs}
The ROI algorithm only requires the raw chromatographic data and two input parameters to find ROIs from GC-MS chromatographic data. Users upload the raw chromatogram, \textbf{data}, as a \textit{I $\times$ J} data matrix and specify an integer value for the moving window size, \textbf{wndw}. The probability threshold, \textbf{CutOff}, also input by the user. This is the minimum probability (between 0 and 1) for a spectrum to be deemed to be within a ROI.

\subsubsection{Algorithm outputs}
The ROI algorithm outputs a series of five vectors of length \textit{I} and one \textit{I $\times$ J} matrix. The vectors include \textbf{pv}, the probabilities for each spectrum being within a ROI, on a scale of 0 to 1; \textbf{modPVans}, where the probabilities in \textbf{pv} above the user-specified threshold \textit{CutOff} are retained, and those below the threshold are set to 0; \textbf{boolCutOff}, a boolean version of \textbf{modPVans}, where probabilites above the threshold are set to 1; \textbf{ticData}, the original total ion chromatogram (TIC) of the input chromatogram; and \textbf{noiseDroppedTIC}, the Hadamard (element-wise) product of \textbf{ticData} and \textbf{boolCutOff}. This last vector, \textbf{noiseDroppedTIC}, contains values of zero outside of identified ROIs, and the original value of the TIC within ROIs. \textit{J} copies of \textbf{boolCutOff} are then stacked to generate a two-dimensional boolean \textit{I $\times$ J} mask that is applied to the original GC-MS chromatogram, \textbf{data}, via another Hadamard product to generate \textbf{noiseDropped}, which is output as the GC-MS data with signals of all ions retained within identified ROIs, and zeros outside of ROIs.

\subsubsection{Region of interest selection}
 ROI identification proceeds as follows: data contained within the moving window with size \textbf{wndw} is first auto-scaled, and then the first two singular values are computed using the svds() function in Matlab. An approximation of a Fisher ratio is computed as the ratio of the squares of the first two singular values. Although we have no formal mathematical proof that this is equivalent to a Fisher ratio, this calculation is similar in form to the calculation of a Fisher ratio. Furthermore the utility of this approach is demonstrated empirically in this work. This process is repeated in each iteration of the algorithm as the window is shifted along the raw data by one spectrum in each iteration. Thusly, a vector of pseudo-F-ratios is generated. From this, a vector of probabilities that each spectrum is within a ROI is generated using the \textit{F} cumulative distribution function (fcdf() in Matlab) with \textbf{wndw}-1 numerator degrees of freedom and \textbf{wndw}-2 denominator degrees of freedom. The degrees of freedom for the numerator was estimated as being the total number of orthogonal vectors, or non-zero singular values, that could possibly be extracted from the rectangular matrix \textit{\textbf{wndw} $\times$ J} assuming that the window size always contains fewer elements than the mass spectral mode. One degree of freedom was removed from the numerator, considering that a vector of means was empirically determined from the data. Since the denominator was calculated on data containing one less possible orthogonal component, its degrees of freedom was estimated as \textbf{wndw}-2.

For each value in the vector of probabilities, the $\chi$\textsuperscript{2} was calculated using the $\chi$\textsuperscript{2}-inverse cumulative distribution function using chi2inv() in Matlab, with 1 degree of freedom and stored as a vector. Vector addition was then performed, combining this vector with the previously summed $\chi$\textsuperscript{2} values, with proper indexing to account for the moving window. The probability for each mass spectrum was then calculated from the $\chi$\textsuperscript{2} values using the chi2cdf() function in Matlab. The number of degrees of freedom for this calculation was the number of instances where the acquisition was considered by the moving window; for instance, the first mass spectrum has one degree of freedom, while a mass spectrum in the middle of a sufficiently large chromatogram (where \textit{n} $>$ 2 $\times$ \textbf{wndw}) will have \textbf{wndw} degrees of freedom. Regions of interest were then selected from the chromatogram by selecting regions of the chromatogram where the probability is higher than the user-input probability threshold. 

\subsection{GC-MS analysis of standards and samples}
GC-MS chromatograms of a standard Grob mix were generated at various concentrations to explore the sensitivity of our ROI algorithm by demonstrating its ability to differentiate chromatographic peaks from baseline noise. Linear saturated alkanes were also analyzed to demonstrate our algorithms ability to handle chromatograms with multiple compounds with high degrees of fragmentation. Three derivatized urine samples were also analyzed to explore how the ROI algorithm handles complex biological samples. 

\subsubsection{Preparation of standards}
A standard Grob mix (Restek, Bellefonte, Pennsylvania, USA) was diluted in dichloromethane (Fisher Chemical\textsuperscript{TM}, Saint-Laurent, Quebec, Canada) to make a 6-point dilution series (30, 15, 7.5, 3, 1.5, and 0.3 $\mu$g/mL). Additionally, starting with a 1000 $\mu$g/mL stock solution of linear, saturated alkanes from C$_7$ to C$_{30}$ in hexane (Sigma-Aldrich, Oakville, Ontario, Canada) a 30 $\mu$g/mL standard was prepared by dilution in hexane (Fisher Chemical\textsuperscript{TM}). Aliquots of 100 $\mu$L of each standard were transferred into clear glass auto-sampler vials with 300 $\mu$L fused inserts and capped with PTFE/silicone septa (Chromatographic Specialties Inc., Brockville, Ontario, Canada). 

\subsubsection{Derivatized urine samples}
Three urine samples were prepared according to the global derivatization of urinary metabolites described by Nam \textit{et al.} (2020) \cite{metabo10090376}. In brief, a fresh urine sample was obtained, then vortexed for 1 minute. Then 40 $\mu$L of urine was aliquoted into a 2 mL centrifuge tube, followed by 20 $\mu$L of internal standard (100 $\mu$g/mL solution of 4-\textsuperscript{13}C methylmalonic acid in water) and 10 $\mu$L of a 40 mg/mL solution of urease in water. The samples were vortexed for three minutes, then incubated at 37 \textsuperscript{$\circ$}C for 1 hour. Next 1860 $\mu$L of methanol (Fisher Chemical\textsuperscript{TM}) was added to each sample, followed by 5 minutes of vortexing and centrifugation for 10 minutes at 10,000 $\times$ g. 1 mL of the supernatant was aliquoted into a 2-mL GC vial (Chromatographic Specialties Inc.), capped with a PTFE/silicone septum then dried under a gentle stream of nitrogen at 50 \textsuperscript{$\circ$}C until dry. Then 50 $\mu$L of 200 mg/mL methoxyamine hydrochloride in pyridine was added to the dried residue and incubated at 60 \textsuperscript{$\circ$}C for 2 hours. Next 100 $\mu$L of MSTFA (N-Methyl-N-(trimethylsilyl)trifluoroacetamide, Sigma-Aldrich) was added and incubated again at 60 \textsuperscript{$\circ$}C for 1 hour. An aliquot of 100 $\mu$L of the derivitized metabolite extract was transferred to a clear glass autosampler vial with 300 $\mu$L fused inserts and capped with a PTFE/silicone septum for GC-MS analysis.   

\subsubsection{GC-MS analysis}
All GC-MS analysis was performed on an Agilent 5975 (Agilent Technologies Inc., Santa Clara, California, USA) GC-MS using a 30 m $\times$ 0.25 mm; 0.25 $\mu$m HP-5MS column (Agilent Technologies Inc.) for separation. A 7683 ALS-GC auto sampler (Agilent Technologies Inc.) was used for automated injections of 1 $\mu$L aliquots of sample. The inlet was held at 250 $^{\circ}$C and a split ratio of 30:1 was used for injection. Separation was performed with helium carrier gas at a constant flow rate of 1 mL/min. The oven was held at 50 $^{\circ}$C for 3 minutes then heated to 325 $^{\circ}$C at a rate of 12.5 $^{\circ}$C/min, then the final temperature was held for 10 minutes. Mass spectra were collected with an acquisition rate of 3.5 spectra/s over a mass range of 35 to 450 m/z. The transfer line temperature was 250 $^{\circ}$C and the ion source temperature was 230 $^{\circ}$C. The total analysis time for each run was 35.0 minutes.

\subsection{Region of interest sensitivity}
To evaluate the ROI algorithm's capability to detect analyte signal at low concentrations, we applied the ROI algorithm to each chromatogram from the Grob mix standard curve, to determine the approximate on-column concentration where the algorithm struggles to differentiate analyte peaks from baseline noise.

\subsection{Flexible Coupling PARAFAC2}
Flexible coupling PARAFAC2 was utilized to extract the mass spectrum from each peak within a region of interest, across all concentrations of Grob mix \cite{cohen2018nonnegative}. For a given ROI, the sub-matrices of the raw chromatographic data contained in the same region for each Grob mix chromatogram were excised and stacked to form an \textit{I $\times$ J $\times$ K} tensor, where \textit{I} is the number of \textit{m/z}, \textit{J} is the number of spectra (corresponding to time), and \textit{K} is the number of Grob mix chromatograms from which the ROI was excised. Thid tensor was then subjected to flexible coupling PARAFAC2. Unique mass spectra ($A_k$) for each compound in each chromatographic segment ($X_k$) were used to assess the performance of the ROI selection tool, against the best possible extracted mass spectrum for each compound as informed by the higher concentration levels (with higher signal-to-noise ratio) in the PARAFAC2 model. This was done via the per-sample component profiles ($B_k$) and relative quantities ($D_k$) (below). This process was applied to each set of matching ROIs across the entire chromatogram.

\begin{equation}\label{eq:parafac2ms}
    A_k = X_k^TB_kD_k(B_k^TD_k^TD_k^TB_k)^{-1}
\end{equation}

Every sample component mass spectrum across each concentration was subjected to identification by mass spectral matching, utilizing the cosine algorithm \cite{stein1994optimization}. The match factor associated with each correctly identified compound was tabulated for each replicate at a given concentration level, with standard error calculated to represent variations in spectral quality between replicates (Table \ref{table:Table1}). Peaks not belonging to standards in the Grob mix were subjected to identification by matching the mass spectrum retrieved from flexible coupling PARAFAC2 on the entire tensor for that region of interest, with the compound returning the highest match factor tabulated in Supplementary Table 1.

\section{Results and discussion}

\subsection{Grob mix and algorithm sensitivity}
\begin{figure}
  \includegraphics[width=\linewidth]{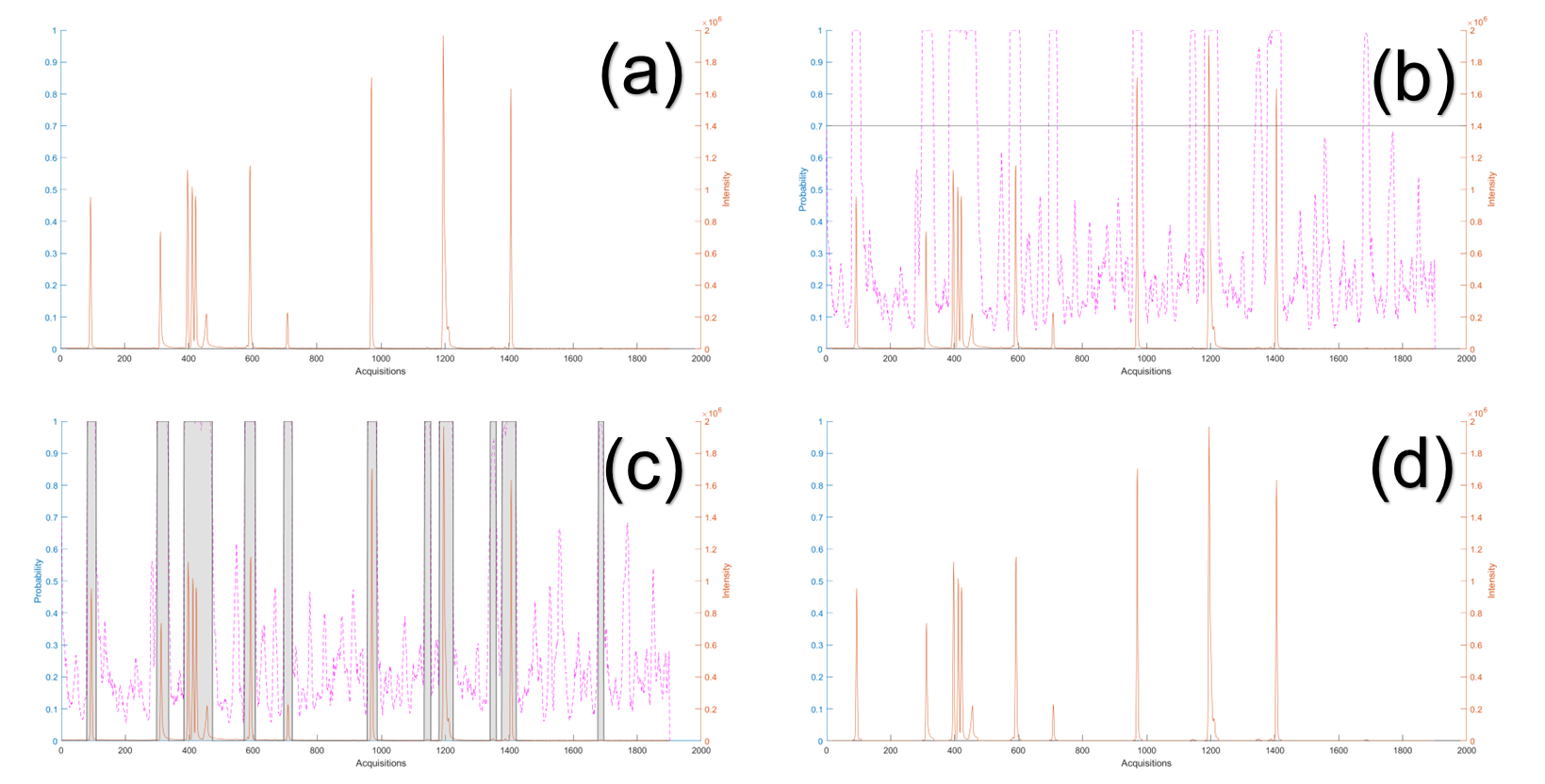}
  \centering
  \caption{Visualization of the key steps in the region of interest selection algorithm. (a) Zoomed in region of a total ion chromatogram (TIC) of Grob mix standard (1000 pg on column) (b) the probability values for each acquisition (dotted line) overlaid on the TIC, probability threshold represented by the horizontal blue line (c) the regions of interest, gray rectangles, superimposed on the TIC (d) chromatographic signal not contained within a region of interest is dropped.}
  \label{fig:figure2}
\end{figure}

ROI algorithm sensitivity was validated by identifying and extracting regions of interest from GC-MS chromatograms of Grob mix. The Grob mix contains a number of compounds which elute closely together, which allowed us to explore how the algorithm handles regions comprising both single, resolved peaks and regions with two or more closely eluting or co-eluting peaks. In figure \ref{fig:figure2}c we see three peaks closely eluting, undecane, nonanal, and 2,6-dimethyl phenol, and they are all encompassed into one ROI. This demonstrates that our algorithm is in fact a region of interest selection tool rather than a peak selection tool. This is ideal for subsequent analysis as deconvolution approaches such as flexible coupling PARAFAC2 perform better with a smaller number of chemical components \cite{cohen2018nonnegative}.

A major advantage of this algorithm is that it only requires two user input parameters for ROI selection, window size and probability cutoff. The probability increases rapidly around a region of interest, likely explaining why we observe little impact of the cut-off on the width of the ROI in terms of spectra (time). However, decreasing the probability cutoff can increase the number of identified ROIs by including peaks in ROIs that have a lower probability of being true peaks. This increase in probability at a region of interest is observed in figures \ref{fig:figure2}a and 1b where the probability near a chromatographic peak jumps rapidly, often up to 1. The noisy regions from the chromatogram (figure \ref{fig:figure2}a) are dropped in figure \ref{fig:figure2}d, by setting all the non-ROI spectra to 0. Setting the non-ROI spectra to 0 is performed to preserve the acquisition number, corresponding to retention time. A probability threshold of 0.7 was selected through a trial and error process for this study; however, our experience suggests that a probability threshold between 0.6 and 0.8 tends to be optimal for most chromatograms.

\begin{figure}
  \includegraphics[width=\linewidth]{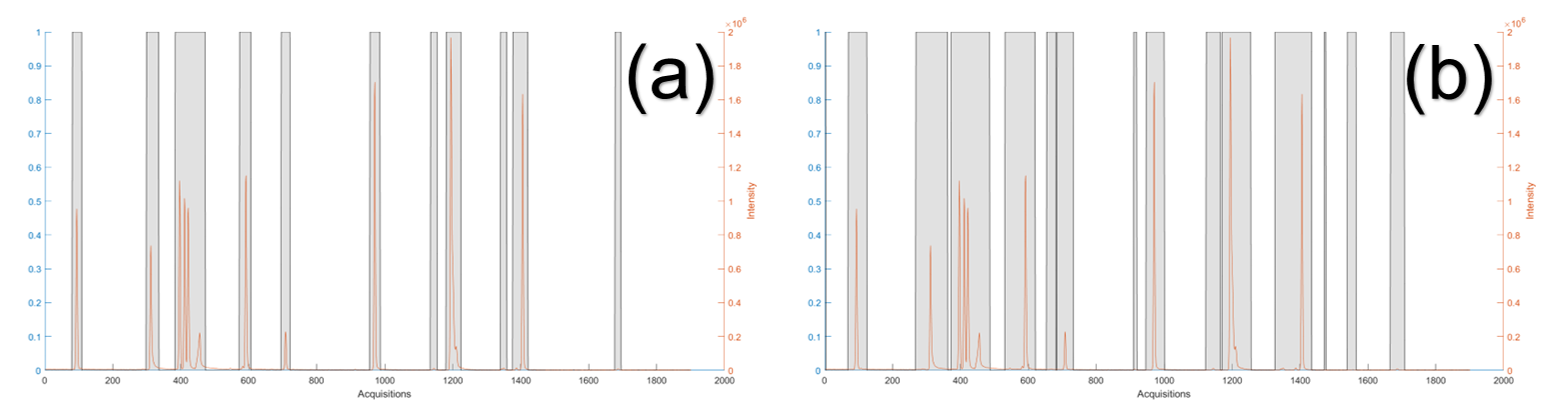}
  \centering
  \caption{The effect of changing the moving window size on a zoomed in chromatogram of Grob mix (1000 pg on column). In (a) the window size used was 10 and (b) a window size of 20 was used. One additional peak is included in a region of interest in (b) at the end of the chromatogram.}
  \label{fig:peakWidths}
\end{figure}

The ROI algorithm is relatively agnostic to moving window size, but we found the ideal window size was approximately the average peak width measured in terms of number of spectral acquisitions. In Figure \ref{fig:peakWidths}, two moving window sizes were explored: 10 spectra, which is approximately the average peak width, and 20 spectra, double the peak width. With a window size of 10 (Figure \ref{fig:peakWidths}a) all the obvious peaks are included in ROIs, and the regions are sufficiently wide to contain the entire peak. Doubling the window size to 20 (Figure \ref{fig:peakWidths}b) has little effect on the ROI quality, other than a slightly increased ROI width around the peaks. An additional peak is included with the larger window size. While this peak appears to be an artifact of contamination due to its low signal-to-noise ratio, it demonstrates the effect that window size has on sensitivity in our algorithm. Both probability cutoff and moving window size require minimal user optimization to obtain informative regions of interest from GC-MS data. 

\begin{figure}
  \includegraphics[width=\linewidth]{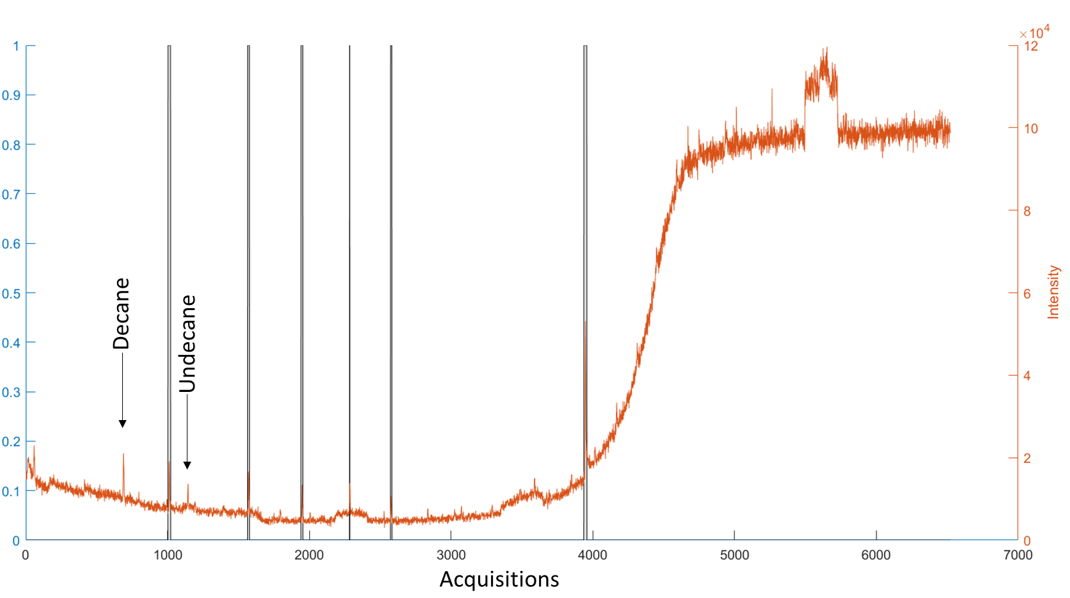}
  \centering
  \caption{Total ion count (TIC) chromatogram of Grob mix at 5 pg on-column, with the selected regions of interest in the gray bars. Regions of interest were selected using a window size of 10 and probability cutoff of 0.7. Decane and undecane peaks are marked to show that while they are above the baseline noise they are below the limit of detection of the ROI algorithm.}
  \label{fig:figure1}
\end{figure}

To evaluate the sensitivity of our ROI selection algorithm, we found the concentration level on-column where the algorithm started to experience peak dropout. Our algorithm had comparable sensitivity to flexible coupling PARAFAC2 decomposition. With the ROI algorithm over the concentration range explored, we only start to see analyte peaks not selected as ROIs at the 5 pg on-column level. In Figure \ref{fig:figure1} at 5 pg on-column, with a window size of 10 and probability cutoff of 0.7, the ROI algorithm does not include decane and undecane in ROIs. Unsurprisingly, the mass spectra recovered from these samples presented a very low match factor as well. This is likely due to the fact that the components themselves are poorly differentiated from the baseline noise. A number of other analytes in figure \ref{fig:figure1} are not detected by the ROI algorithm as their mass spectral quality is too poor to be differentiated from the baseline. This is not a failure of the algorithm, but is more just a fundamental limitation on any type of interpretation of data close to the baseline noise level.

The match factors calculated with the cosine algorithm for each analyte's mass spectrum extracted by flexible coupling PARAFAC2 deconvolution at all seven concentrations of Grob mix are given in table \ref{table:Table1} \cite{stein1994optimization}. This method for identifying compounds experiences sensitivity issues around 10 pg on-column for three of the standards, then at 5 pg on column about half of the standards are not correctly identified from their mass spectra. One analyte, 2,3-butanediol, was absent from our analysis, even though it is present in the Grob mix. The Van Den Dool and Kratz retention index for 2,3-butanediol on a HP-5ms column is 819 \cite{https://doi.org/10.1002/jsfa.845}. In the chromatogram of the linear saturated alkane standards mix (C$_7$ to C$_{30}$) the first alkane to appear after the solvent delay was nonane, confirmed by mass spectral matching. Both heptane and octane do not appear in the alkane chromatograms (Figure \ref{fig:alkaneFigure}), therefore we suspect that 2,3-butanediol, with a similar retention index to octane, is eluting during the solvent delay.

\begin{table}[]
\small
\caption{Match factors for each analyte in the Grob mix identified in the chromatogram across the 7 concentration ranges, expressed as pg on column. Error is given as standard error, where N = 3, and $t_R$ is given in minutes. Note: * denotes N = 2 and ** denotes N = 1.}
\centerline{}
\makebox[\linewidth]{
\begin{tabular}{lcccccccc}
\hline
\textbf{Compound} & \textbf{$t_R$} & \textbf{1000 pg} & \textbf{500 pg} & \textbf{250 pg} & \textbf{100 pg} & \textbf{50 pg} & \textbf{10 pg} & \textbf{5 pg} \\ \hline
Decane               & 6.8  & 97.7±0.1 & 97.8±0.1 & 97.9±0.1 & 97.8±0.1 & 97.7±0.3 & 86.6±1.8  & 76.2**    \\
1-Octanol            & 7.9  & 88.8±0.1 & 88.4±0.2 & 88.1±0.2 & 86.7±0.7 & 86.8±0.6 & 68.2±0.2* & ND        \\
Undecane             & 8.3  & 96.9±0   & 96.9±0   & 96.7±0.1 & 96.6±0.1 & 96±0.3   & 84.9±1.3  & 80.9±4.6  \\
Nonanal              & 8.4  & 69.7±2.4 & 71±1.2   & 68.7±1.3 & 68.7±1.8 & 71.5±3.6 & 71.2±2.9  & 66.9±4.3  \\
2,6-dimethyl phenol  & 8.5  & 99±0     & 99.1±0   & 99.1±0   & 99.1±0   & 99±0.1   & 92.5±0.5  & 73.3±1.6  \\
2-ethylhexanoic acid & 8.6  & 72.5±2.4 & 73.4±1.3 & 70.8±0.2 & 64±1.4   & 62.2±3   & ND        & ND        \\
2,6-dimethyl aniline & 9.2  & 99.7±0   & 99.6±0   & 99.7±0   & 99.7±0.1 & 99.3±0.2 & 95.6±0.7  & 93.3±0.6  \\
Methyl Decanoate     & 11.1 & 98.4±0.1 & 98.1±0.1 & 98.5±0.1 & 98.2±0.2 & 97.6±0.2 & 69.8±4.1  & 46**      \\
Dicyclohexylamine    & 12.1 & 99.8±0   & 99.8±0   & 99.8±0   & 99.6±0.1 & 98.8±0.3 & 85.2**    & 93.4±3*   \\
Methyl undecanoate   & 13.1 & 83.4±0.1 & 83.5±0.1 & 83.5±0   & 83.6±0.1 & 83.4±0   & 71.4±3.1  & 68.6±1*   \\
Methyl laurate       & 15.9 & 91.6±0   & 91.3±0.2 & 91.6±0.1 & 91.5±0.1 & 90.8±0.8 & 80.8±1.4  & 66.6±10.6 \\ \hline
\end{tabular}
}
\label{table:Table1}
\end{table}

The ROI algorithm failed at approximately 5 pg on column, which is where the mass spectra extracted with flexible coupling PARAFAC2 were also no longer identifiable (Table \ref{table:Table1}). It is important to note that the ROI algorithm is agnostic to fluctuations in the baseline, as seen in the column bleed section of the chromatogram in Figure \ref{fig:figure1}. This is because our algorithm is looking for differences in the mass spectra contained in the moving window. Baseline fluctuations are the result of an increase in baseline noise, and while the mass spectral signal of the baseline might increase, the actual spectral information does not change. Therefore our algorithm does not pick up significant baseline shifts. In addition to the 11 compounds in the Grob mix detected by the ROI algorithm, 10 additional peaks are selected as ROIs. To confirm their identity, flexible coupling PARAFAC2 was used to extract their mass spectra, with putative identification by spectral matching. They were identified as relatively obscure compounds, confirming they are the result of contamination in the GC-MS system.

\subsection{Region of interest selection of linear alkanes}

\begin{figure}
  \includegraphics[width=\linewidth]{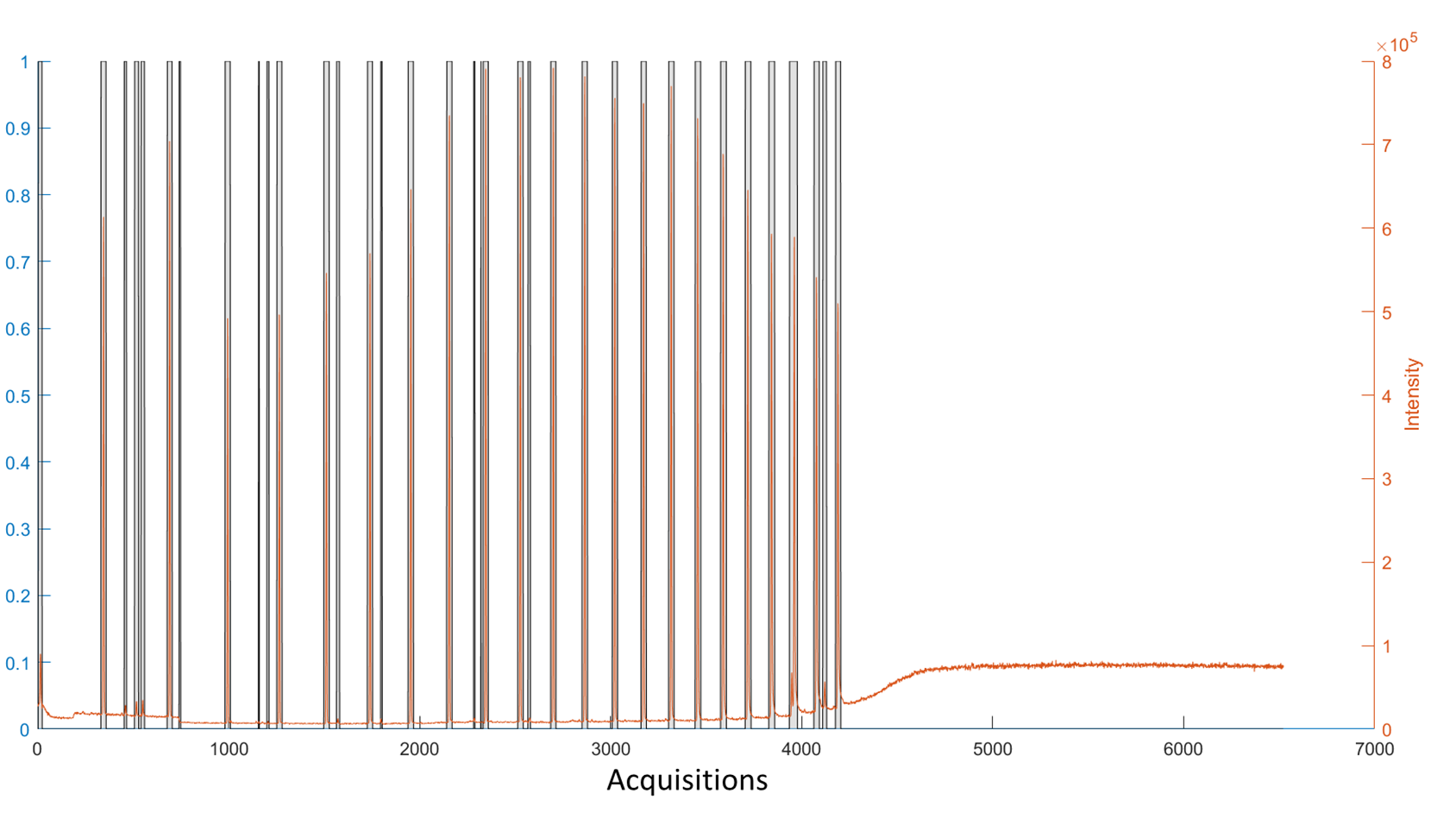}
  \centering
  \caption{Chromatogram of saturated linear alkanes, C$_9$ to C$_{30}$, at 1000 pg on column with regions of interest superimposed on the chromatogram (probability cutoff of 0.7, window size of 10).}
  \label{fig:alkaneFigure}
\end{figure}

Our algorithm also performs well with selecting ROIs from a chromatogram of linear saturated alkanes. Linear alkanes undergo a high degree of fragmentation with EI ionization, making their mass spectra more challenging to distinguish from baseline noise. We can detect every compound in the alkane mix, without peak dropout or peak splitting (Figure \ref{fig:alkaneFigure}). As seen with the Grob mix, our ROI algorithm is agnostic to shifts in the baseline seen in Figure \ref{fig:alkaneFigure}. A number of additional low-intensity peaks are detected in Figure \ref{fig:alkaneFigure} by the algorithm. The mass spectra of these peaks are too low quality to perform putative identification, and we suspect they are contaminants in the GC-MS system. Nonetheless, the detection of these peaks that are orders of magnitude less intense than the neighbouring alkanes demonstrates the ability of our algorithm to detect peaks across a wide range of intensities within the same chromatogram.

\subsection{Derivatised urine region of interest selection}
\begin{figure}
  \includegraphics[width=\linewidth]{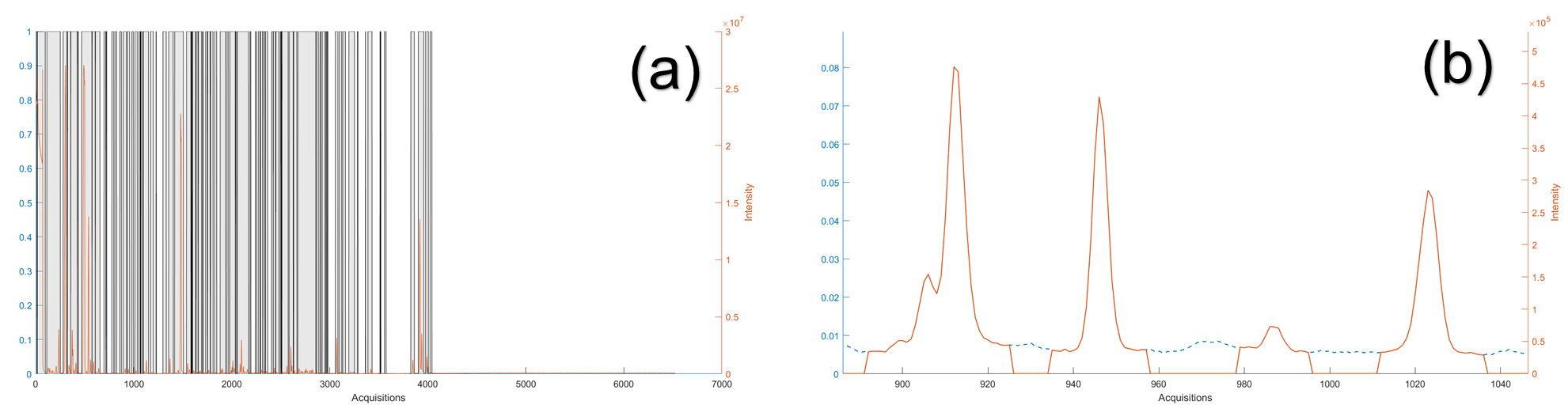}
  \centering
  \caption{(a) Total ion chromatogram (TIC) of derivatised urine with regions of interest overlaid. Regions were selected using a window size of 10 and probability cutoff of 0.7. (b) Enlarged region of the urine chromatogram TIC with the ROI portions (solid) and noise (dashed) components of the chromatogram shown.}
  \label{fig:urineChr}
\end{figure}

The algorithm performs well on the complex derivatized urine samples, with all peaks being contained within ROIs (Figure \ref{fig:urineChr}a). As there are hundreds of unique chemical compounds present in the derivatized urine matrix, many of the peaks are not well resolved, unlike the Grob and alkane standards, therefore a number of ROIs contain multiple peaks. Minimal trial and error was required to find the ideal window size and probability cutoff (10 and 0.7, respectively), demonstrating the robustness and simplicity of our algorithm when handling complex chromatograms. A window size of 5 and 15 were explored on the urine chromatogram (Supplemental Figure 9). The smaller window size experienced significant dropout of ROIs, while the larger window size had slightly larger ROIs, encompassing more noise than the window size of 10. Our ROI algorithm detects peaks in the urine chromatogram over a large dynamic range of intensities. Figure \ref{fig:urineChr}b demonstrates the successful differentiation between chromatographic signal and noise by our algorithm. In the enlarged section of the chromatogram, containing 4 ROIs, the detected small, low-intensity peaks are in fact real chemical features and not baseline fluctuations.

It is important to note that our algorithm is not a peak detection algorithm, as it is not looking for individual chromatographic peaks, but is instead looking for regions in the data where there is chromatographic information that is distinguishable from the baseline. This is best represented in figure \ref{fig:urineChr}c where multiple peaks are encompassed in a single ROI. A particular ROI can then be subjected to further chemometric techniques such as MCR, PARAFAC or PARAFAC2.

\section{Conclusion}
This work reveals the application of pseudo-Fisher ratio analysis for sensitive, untargeted region of interest selection in GC-MS data. Informative ROIs are generated with only two user input parameters, moving window size and probability cutoff, therefore minimal user trial and error is required. Our algorithm detects regions of interest over a large dynamic range, and only begins to experience dropout at low concentrations when the mass spectral quality of peaks within a region becomes indistinguishable from the baseline noise. This approach is also agnostic to major fluctuations in baseline noise within a chromatogram due to the scaling of the data prior to pseudo-Fisher ratio analysis. Complex samples, such as the derivatized urine explored here, are efficiently handled by the algorithm, successfully incorporating all chromatographic peaks into regions of interest. The data generated by our algorithm is easily extracted as a data matrix for further analysis. Our approach demonstrates an effective component in a workflow for fully automated data analysis of GC-MS data.

\section{Acknowledgements}
The authors would like to thank Robin Abel from MBC Group in Edmonton, Alberta for their assistance with data curation and consultation. As well we would like to thank Dr. Randy Whittal and Jing Zheng from the University of Alberta mass spectrometry facility for their assistance with the GC-MS system.

\section{Supplementary Information}
The raw GC-MS data sets used in this study are available at \url{https://doi.org/10.5683/SP3/3OEMJY}. The table of putative identities assigned to unknown peaks in the Grob mix chromatogram is available as Table S1 in the supporting information. Additionally chromatograms of Grob mix with overlaid regions of interest across the 7 concentrations explored are available. The source code for this project is available on GitHub at \url{https://github.com/ryland-chem/RegionOfInterest}.

\section{Author Contributions}
Conceptualization, M.D.S.A.; Methodology, R.T.G. and M.D.S.A.; Software development, R.T.G. and M.D.S.A.; GC-MS data curation, R.T.G.; Testing and validation, R.T.G. and M.D.S.A. and A.P.d.l.M; Resources, J.J.H.; Writing-original draft preparations, R.T.G. and M.D.S.A.; Writing-review and editing, R.T.G. and M.D.S.A. and A.P.d.l.M and J.J.H.; Supervision, J.J.H.; project administration, J.J.H.; Funding acquisition, J.J.H. All authors have read and agreed to the published version of the manuscript.

\section{Funding}
The authors wish to acknowledge the support of this research from Genome Canada and Genome Alberta through their Genome Technology Platform (GTP) grant, and the Canada Foundation for Innovation Major Science Initiatives (CFI-MSI) grant to The Metabolomics Innovation Centre, TMIC. The support of the Natural Sciences and Engineering Research Council of Canada (NSERC) through a scholarship to RTG and grants to JJH is also acknowledged.

\section{Conflict of Interest}
The authors declare no conflict of interest.

\bibliography{mybibfile}

\pagebreak
\section{Graphical Abstract}

\begin{center}
  \includegraphics[width=130mm,scale=0.5]{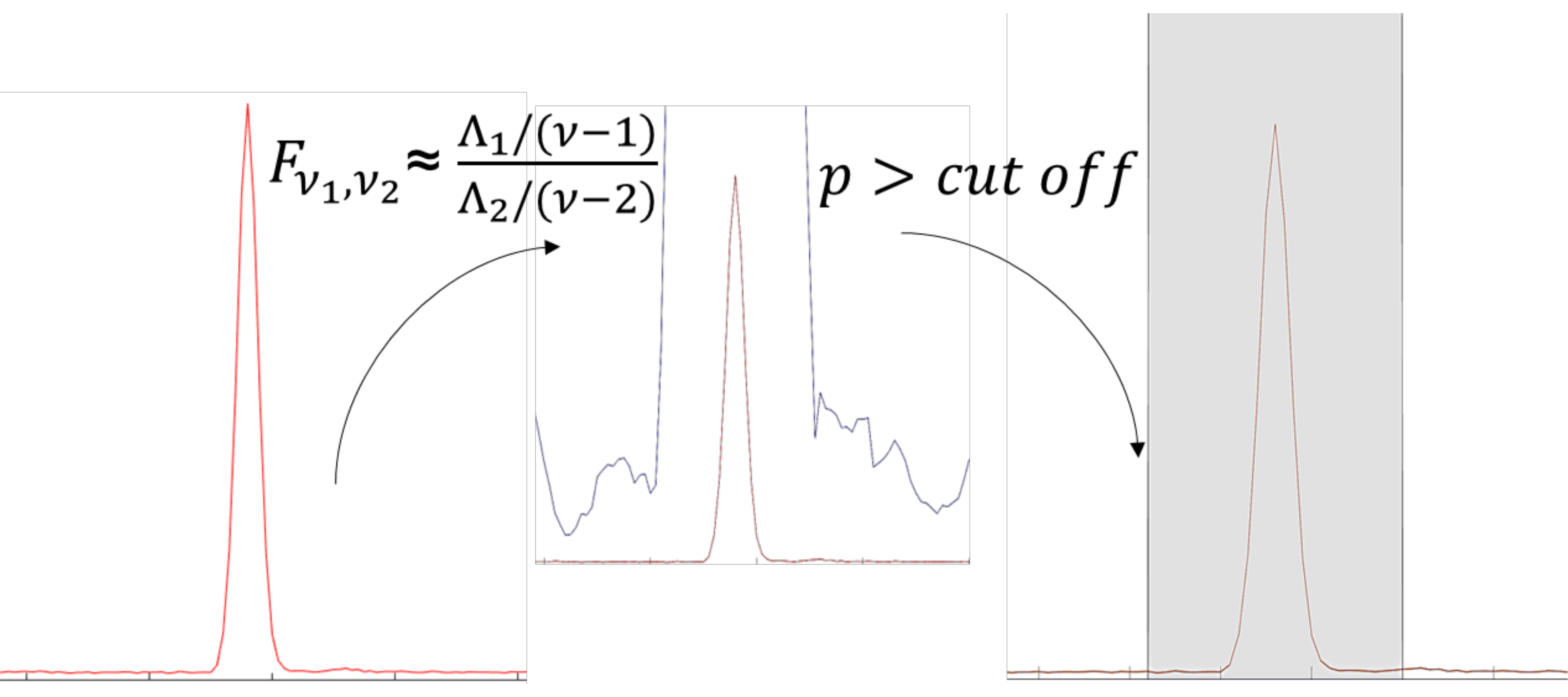}
\end{center}

\end{document}


\maketitle 

\thispagestyle{empty}
\listoftables
\listoffigures
\newpage
\pagenumbering{arabic}

\section{Mathematical Motivation}

It is an extremely difficult task to predict the eigenvalue distribution of random matrices, much less similar distributions of matrices with structured information. Nonetheless, we have chosen to interpret the eigenvalues of a matrix as variances, whose ratios are proportional to the ratio of two independent $\chi^2$ values, with degrees of freedom related to the window size minus 1 for the first eigenvalue, and the window size minus 2 for the second eigenvalue (assuming the data is auto-scaled). Making these assumptions, the algorithm exploits the following relationship related to the $F$-Distribution:

\begin{equation}
    F(\nu_1,\nu_2) = \frac{\chi_1^2 / \nu_1}{\chi_2^2 / \nu_2}
\end{equation}

Which is a generalization of the $F$-distribution that does not require explicit class labels. An additional requirement for this relationship to hold is that the two $\chi^2$ values be statistically independent. To this end, the orthogonality of each eigenvalue's corresponding eigenvector relative to the other are assumed to be sufficiently independent.

\begin{figure}
    \centering
    \includegraphics[width=\textwidth]{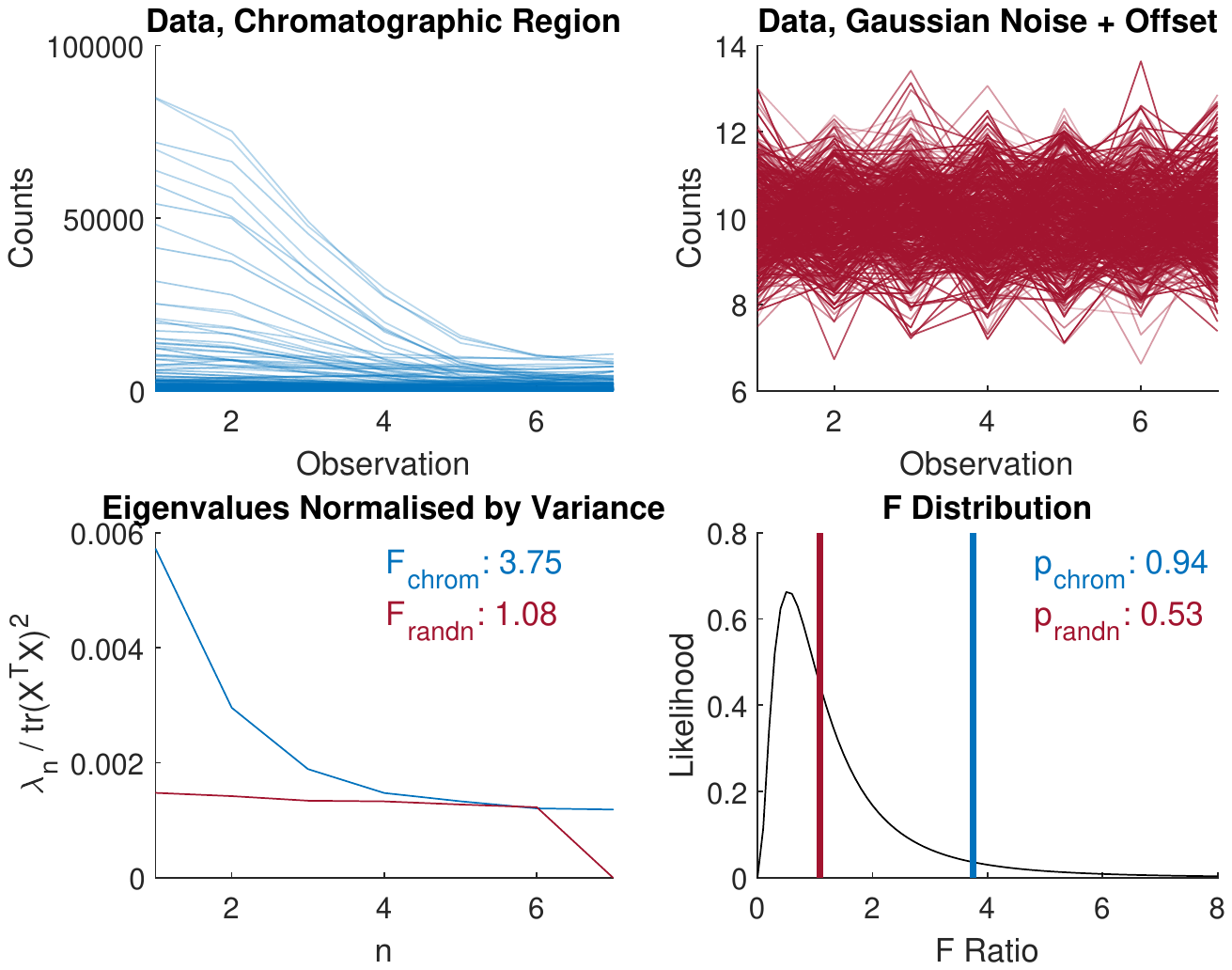}
    \caption{Comparison of pseudo $F$ ratios for matrices containing chromatographic information, versus random Gaussian noise. Regions containing chromatographic information score more highly on an equivalent $F$ distribution according to their pseudo $F$ ratio.}
    \label{fig:fratio}
\end{figure}

It would be incorrect to claim that the distribution of eigenvalues for matrices containing structured information, as is commonly encountered for hyphenated, chromatographic data, is $F$ distributed since by definition the first eigenvalue must always describe an axis of variance that is greater than the second eigenvalue. Therefore all $F$ ratios less than 1 for eigenvalues are not observable. With this in mind; however, matrices that contain structured information (such as a chromatographic peak), versus random noise will typically present a primary eigenvalue that describes this more significant axis of variance. While there is no guarantee that the first eigenvalue will always describe the chromatographic information, and the second eigenvalue will always describe the noise, it is safe to assume that the primary axis of variance increases relative to the secondary axis of variance in those regions of the chromatogram where a chromatographic peak is present. This is reflected in the ratio of the two eigenvalues, which can be ascribed a probability that scales with its position along the $F$ distribution according to the corresponding cumulative distribution function (Figure \ref{fig:fratio}. For this reason, this approximation is useful, but not wholly representative. Hence, its designation as a ``pseudo'' $F$-ratio. 

\section{Evaluation on Grob Mix Standards}
 
 \begin{table}[]
\small
\caption{Retention times (min), top match, and match factors for peaks appearing in the summed Grob standards chromatogram that did not correspond to the 11 known compounds in the standards mix.}
\centerline{}
\makebox[\linewidth]{
\begin{tabular}{ccc}
\hline
\textbf{$t_r$} & \textbf{Top Match}                                    & \textbf{Match Factor} \\ \hline
3.6            & Benzenamine, N-(3-phenyl-2-propenylidene)-            & 86.9                  \\
9.0              & 4-Phosphaspiro[2.4]hept-5-ene, 4-oxo-4,5,6-triphenyl- & 65.8                  \\
9.8            & Oxalic acid, 2-isopropylphenyl octadecyl ester        & 58.9                  \\
11.9           & N-Desmethyltapentadol                                 & 46.8                  \\
12.9           & 2-(2-oxo-1,3,4-oxathiazol-5-yl)phenyl cinnamate       & 65.5                  \\
14.5           & 7-Nitro-1-tetralone oxime acetate                     & 72.6                  \\
16.2           & 2-Chloroaniline-5-sulfonic acid                       & 57.8                  \\
16.5           & 7-Nitro-1-tetralone oxime acetate                     & 78.8                  \\
17.0             & 7-Nitro-1-tetralone oxime acetate                     & 79.8                  \\
22.5           & 9-Octadecenamide                                      & 85.0                    \\ \hline
\end{tabular}
}
\label{table:SupTable1}
\end{table}
 
  \begin{figure}
  \includegraphics[width=\linewidth]{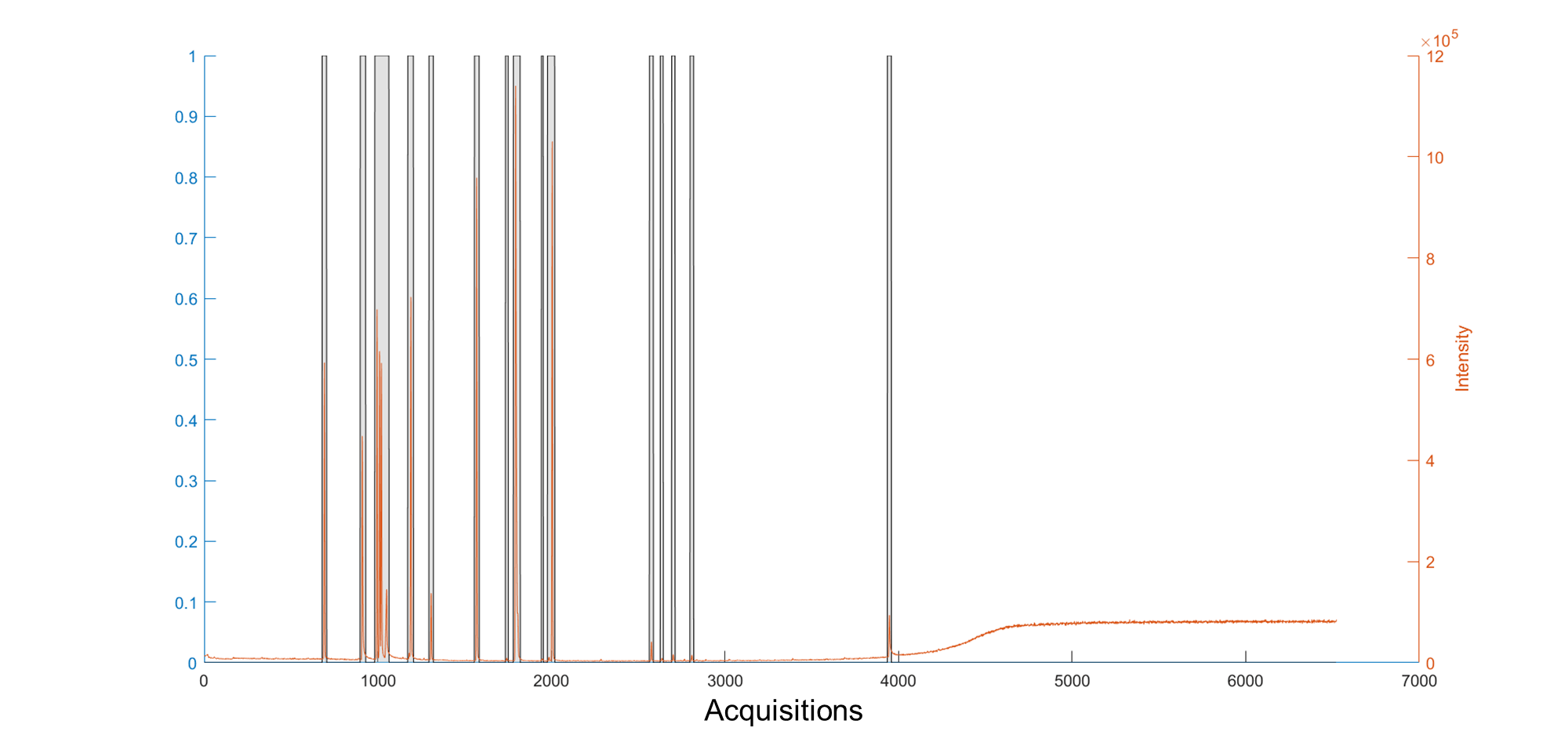}
  \centering
  \caption{Total ion chromatogram of Grob mix at a concentration of 1000 pg on column, with superimposed regions of interest, using a window size of 10 and probability cutoff of 0.7.}
  \label{fig:1000pgOnCol}
\end{figure}

  \begin{figure}
  \includegraphics[width=\linewidth]{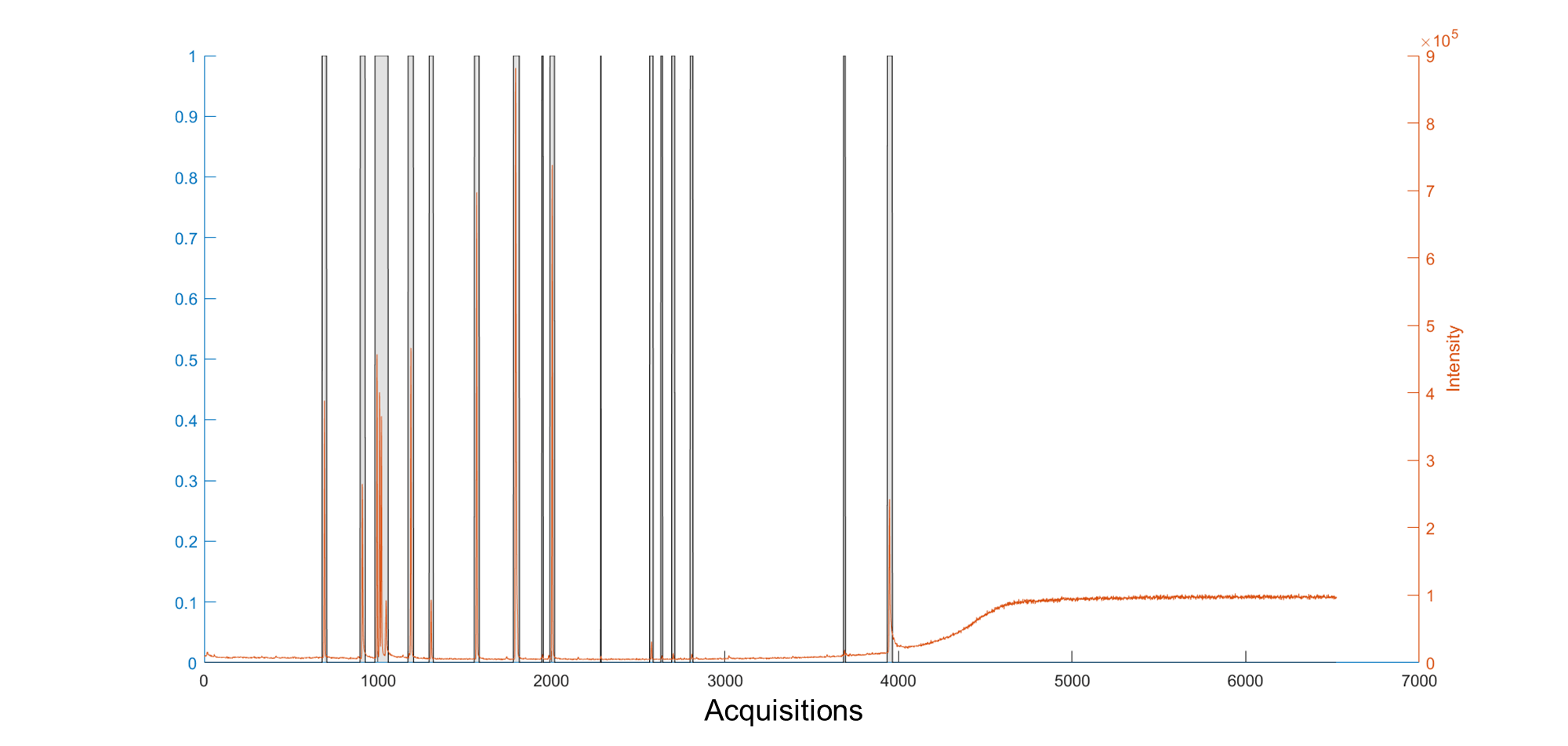}
  \centering
  \caption{Total ion chromatogram of Grob mix at a concentration of 500 pg on column, with superimposed regions of interest, using a window size of 10 and probability cutoff of 0.7.}
  \label{fig:500pgOnCol}
\end{figure}
 
  \begin{figure}
  \includegraphics[width=\linewidth]{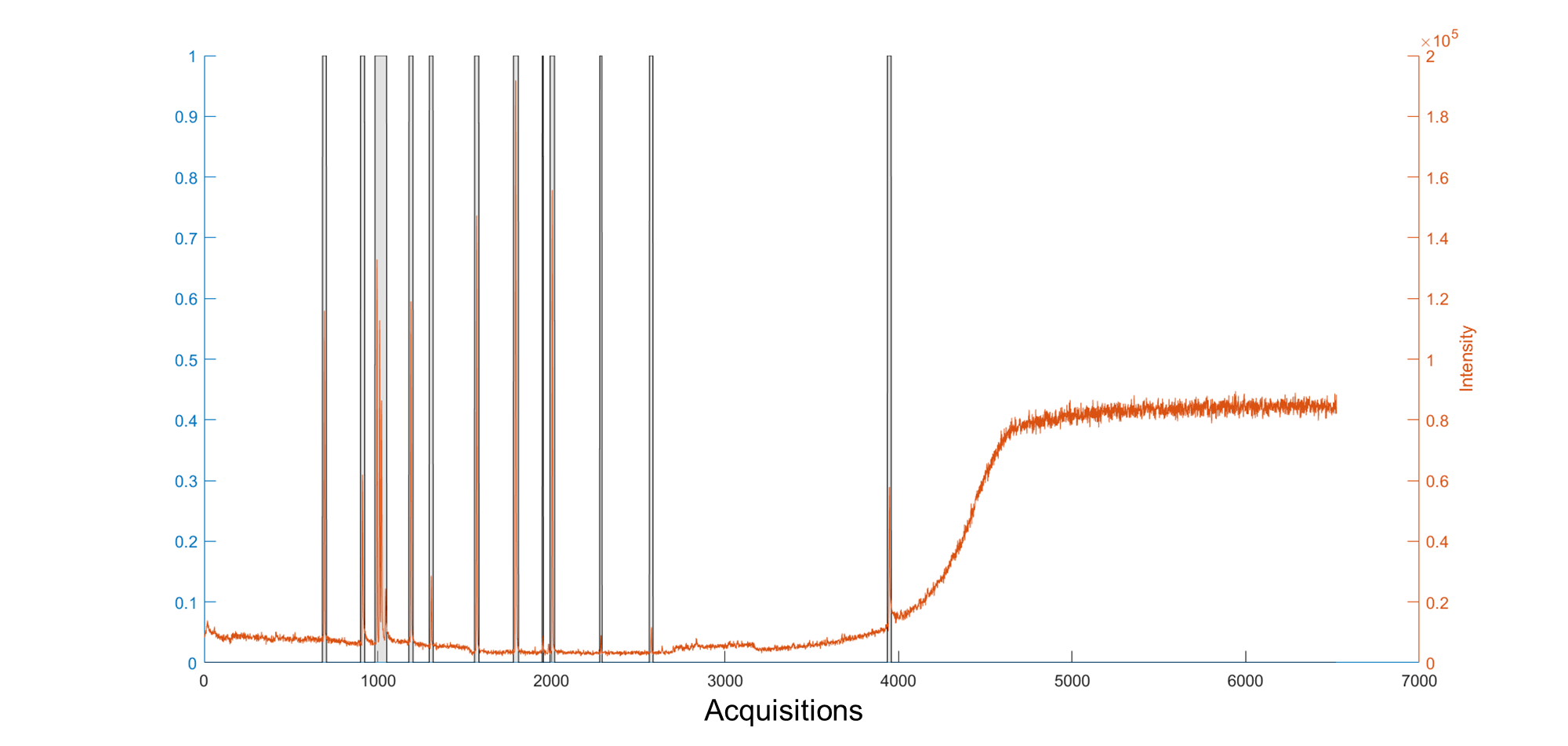}
  \centering
  \caption{Total ion chromatogram of Grob mix at a concentration of 250 pg on column, with superimposed regions of interest, using a window size of 10 and probability cutoff of 0.7.}
  \label{fig:250pgOnCol}
\end{figure}
 
  \begin{figure}
  \includegraphics[width=\linewidth]{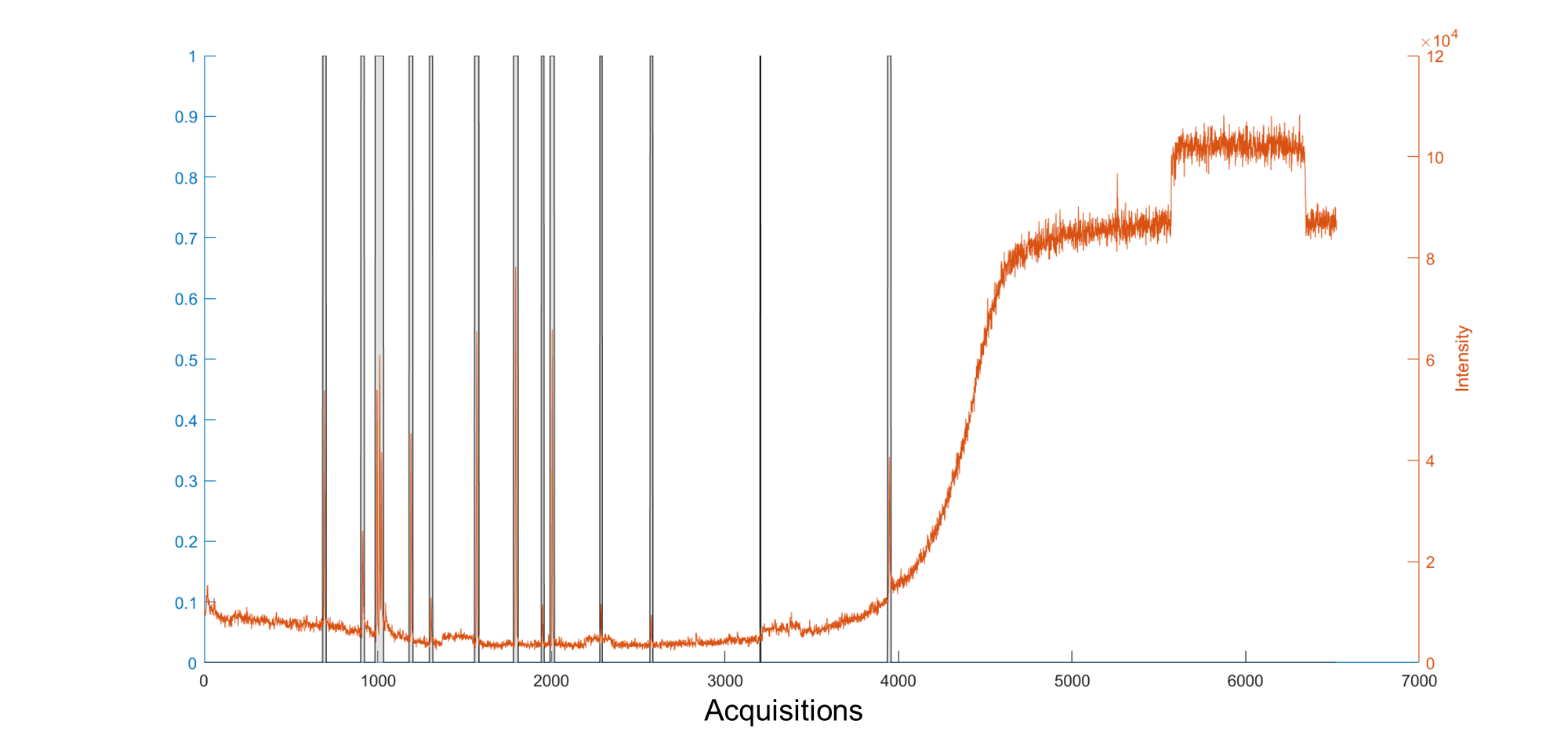}
  \centering
  \caption{Total ion chromatogram of Grob mix at a concentration of 100 pg on column, with superimposed regions of interest, using a window size of 10 and probability cutoff of 0.7.}
  \label{fig:100pgOnCol}
\end{figure}
 
  \begin{figure}
  \includegraphics[width=\linewidth]{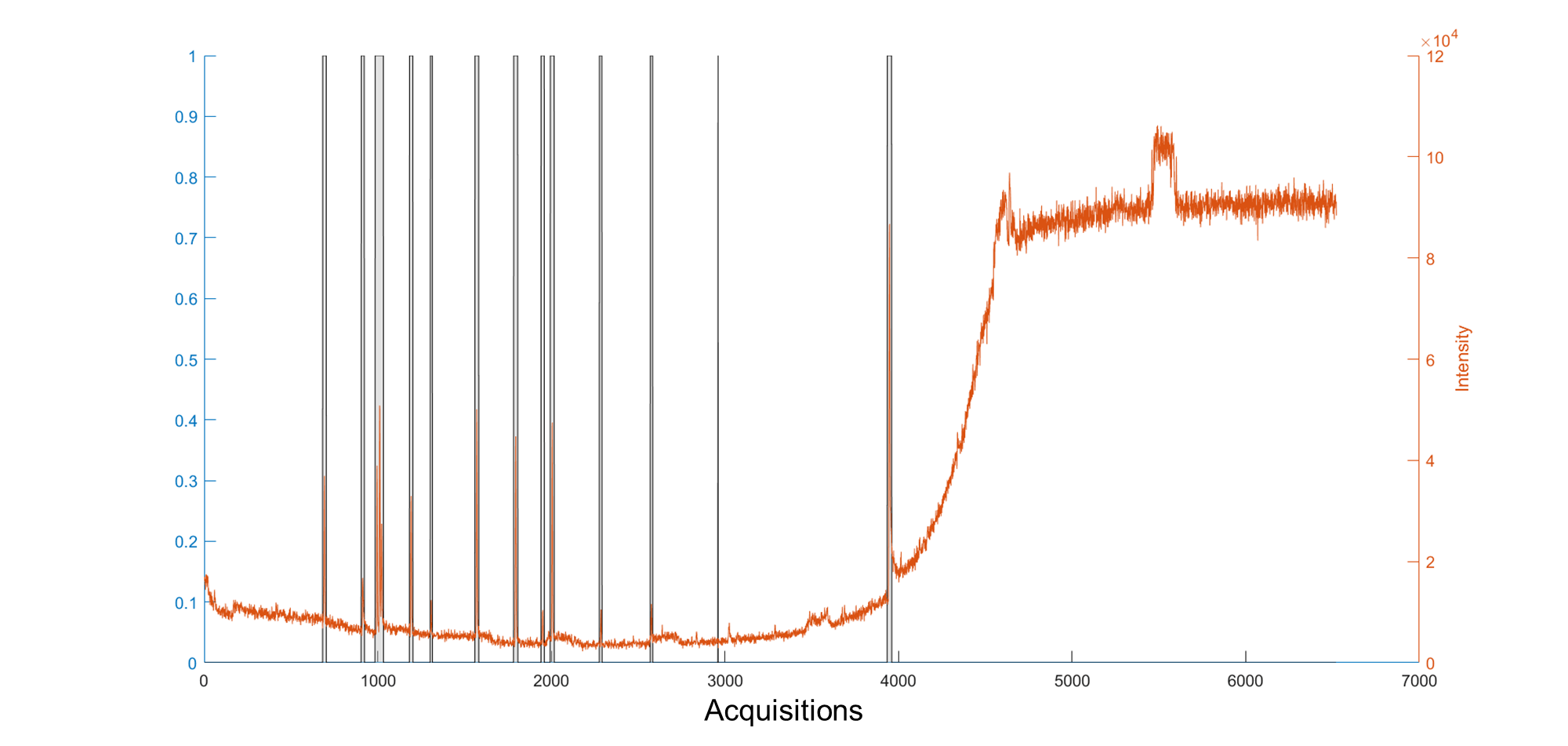}
  \centering
  \caption{Total ion chromatogram of Grob mix at a concentration of 50 pg on column, with superimposed regions of interest, using a window size of 10 and probability cutoff of 0.7.}
  \label{fig:50pgOnCol}
\end{figure}
 
  \begin{figure}
  \includegraphics[width=\linewidth]{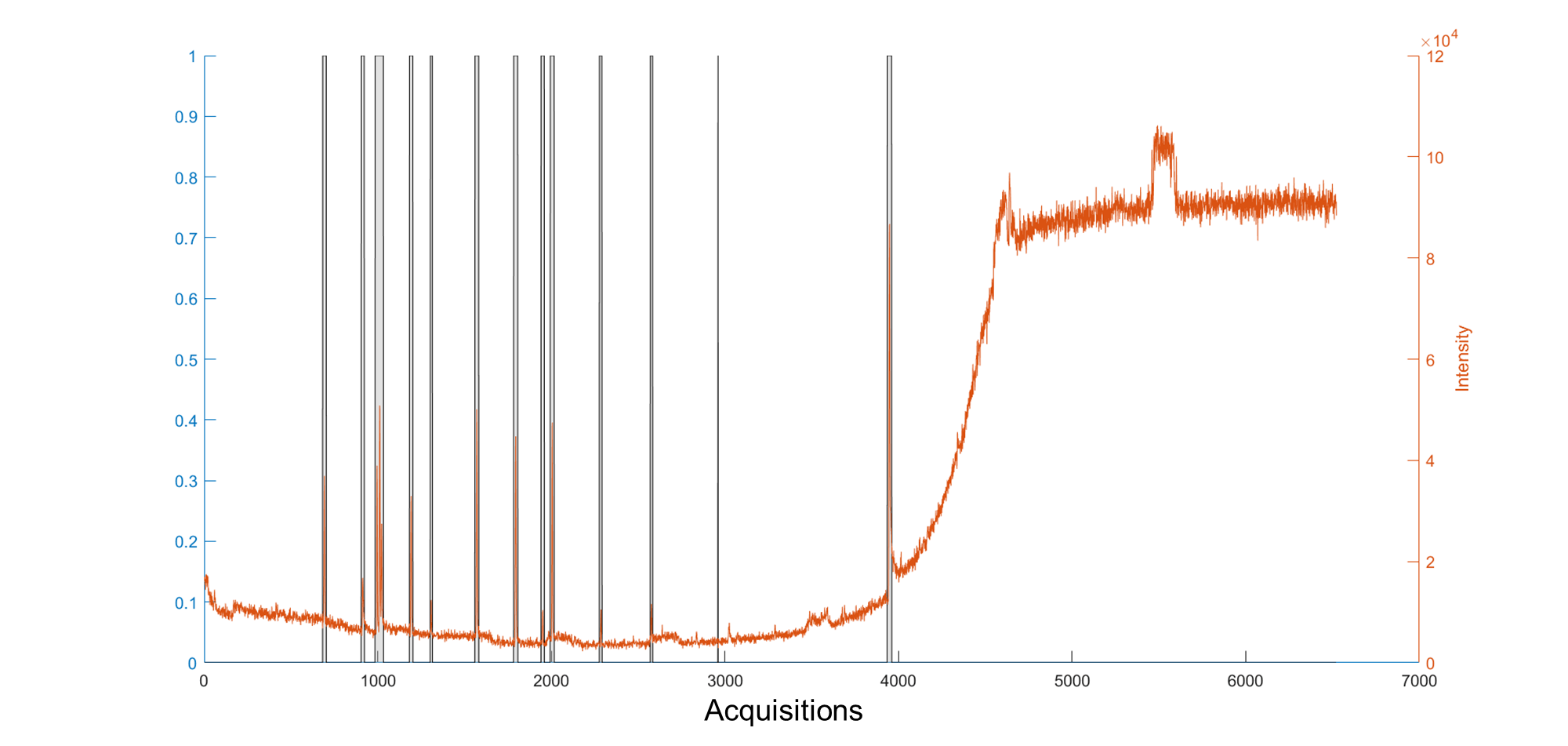}
  \centering
  \caption{Total ion chromatogram of Grob mix at a concentration of 10 pg on column, with superimposed regions of interest, using a window size of 10 and probability cutoff of 0.7.}
  \label{fig:05pgOnCol}
\end{figure}
 
 \begin{figure}
  \includegraphics[width=\linewidth]{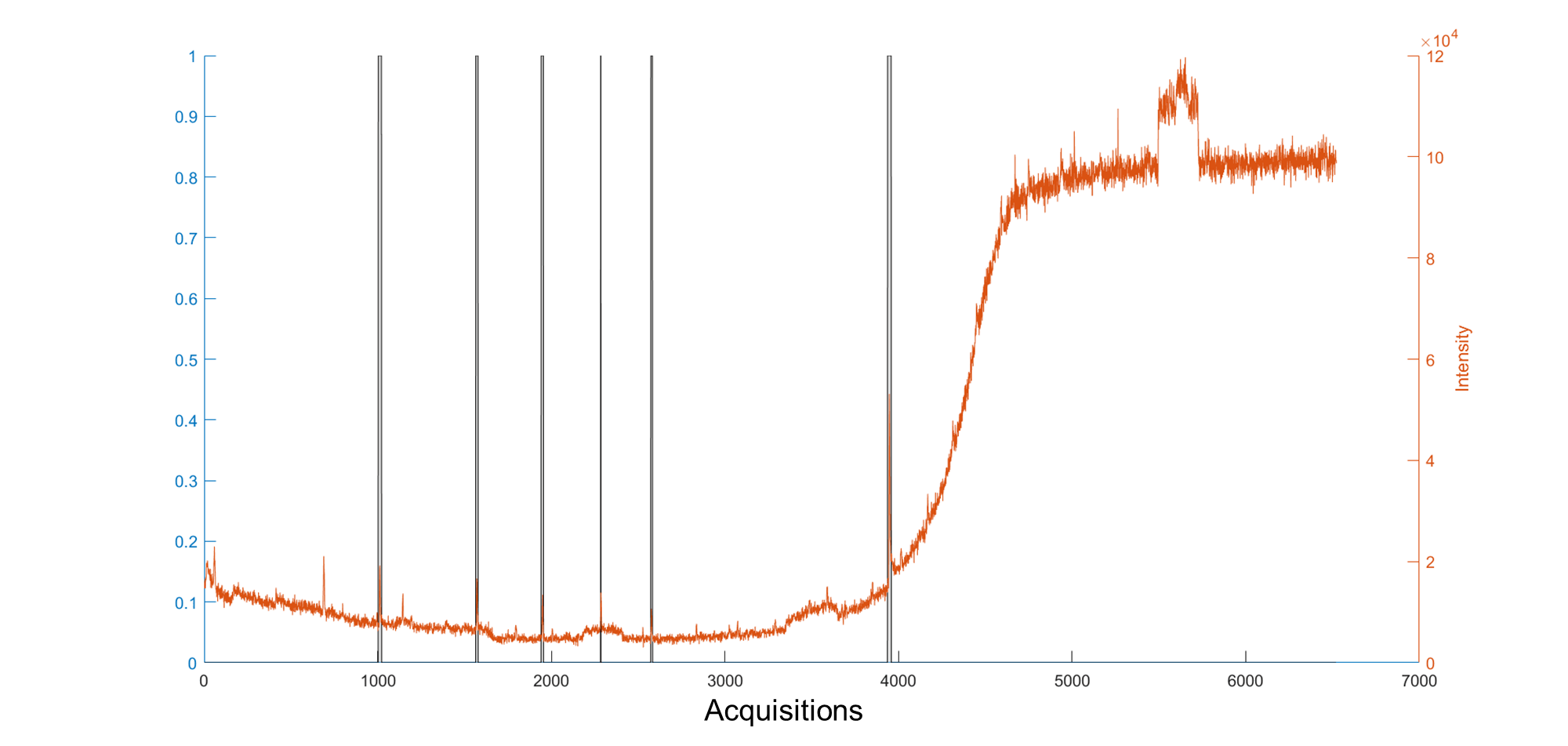}
  \centering
  \caption{Total ion chromatogram of Grob mix at a concentration of 5 pg on column, with superimposed regions of interest, using a window size of 10 and probability cutoff of 0.7.}
  \label{fig:5pgOnCol}
\end{figure}

 \begin{figure}
  \includegraphics[width=\linewidth]{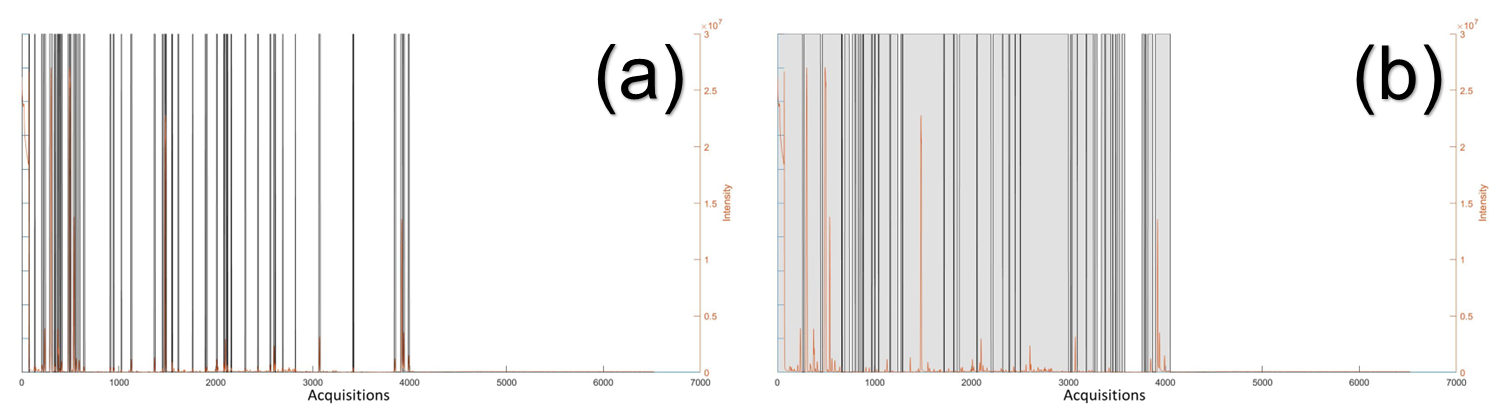}
  \centering
  \caption{Total ion chromatograms of derivatized urine with superimposed regions of interest, using a probability cutoff of 0.7 and a moving window size of 5 (a) and 15 (b).}
  \label{fig:urine_windows}
\end{figure}